\pdfoutput=1
\documentclass[conference]{IEEEtran}
\IEEEoverridecommandlockouts
\usepackage{cite}
\usepackage{amsmath,amssymb,amsfonts}
\usepackage{algorithmic}
\usepackage{graphicx}
\usepackage{textcomp}
\usepackage{xcolor}
\usepackage{comment}

\usepackage{booktabs}
\usepackage{multirow}
\usepackage{colortbl}
\usepackage{ulem}

\usepackage{adjustbox}
\usepackage{array}

\usepackage{rotating}
\usepackage{float}

\usepackage[most]{tcolorbox}
\usepackage{fontawesome5}

\definecolor{warnred}{RGB}{220,38,38}

\newtcolorbox{paperwarningbox}{
  colback=white,
  colframe=warnred,
  boxrule=0.9pt,
  arc=0pt,
  left=8pt,
  right=8pt,
  top=6pt,
  bottom=6pt,
  width=\linewidth
}

\usepackage[colorlinks=true, linkcolor=blue, citecolor=blue, urlcolor=blue]{hyperref}  

\def\BibTeX{{\rm B\kern-.05em{\sc i\kern-.025em b}\kern-.08em
    T\kern-.1667em\lower.7ex\hbox{E}\kern-.125emX}}

\definecolor{headergray}{gray}{0.92}
\definecolor{rowlight}{RGB}{245,245,245}
\definecolor{rowdark}{RGB}{235,235,235}

\definecolor{bestblack}{RGB}{0,0,0}
\definecolor{secondblack}{RGB}{0,0,0}

\newcommand{\cmark}{\textcolor{bestblack}{\checkmark}}
\newcommand{\xmark}{\textcolor{black}{\texttimes}}
\newcommand{\best}[1]{\textcolor{bestblack}{\textbf{#1}}}
\newcommand{\second}[1]{\textcolor{secondblack}{\underline{#1}}}

\begin{document}

\title{META-IR: Meta-Introspective Reasoning for Adaptive Safety Alignment in LLM}

\author{
Mahabubur Rahman Miraj\textsuperscript{1},
Hongyu Huang\textsuperscript{1*},
Nankun Mu\textsuperscript{1},
Chao Chen\textsuperscript{1},
Shanqin Yi\textsuperscript{2}\\
\textsuperscript{1}College of Computer Science, Chongqing University, Chongqing, China\\
\textsuperscript{2}School of Economics and Business Administration, Chongqing University, Chongqing, China\\
\texttt{miraj@stu.cqu.edu.cn, hyhuang@cqu.edu.cn, nankun.mu@cqu.edu.cn,} \\
\texttt{cschaochen@cqu.edu.cn, shanqinyi@stu.cqu.edu.cn} \\
\textit{*Corresponding author: Hongyu Huang}
}

\maketitle

\begin{abstract}
Large Language Models (LLMs) remain vulnerable against jailbreak attacks and harmful-content prompts, particularly under adversarial distribution shifts where static alignment pipelines fail to generalize. Although introspective methods such as STAIR strengthen refusal consistency, they depend on fixed self-evaluation policies that cannot rapidly adapt to evolving manipulation strategies. To overcome these limitations this paper introduces META-IR, a meta-introspective reasoning framework that enhances safety alignment through meta-learning–driven adaptation. Its employs a dual-loop architecture: an inner introspective loop that performs structured risk analysis, self-assessment, and reflective correction via SI-MCTS, and an outer meta-learning loop that optimizes the introspection policy across diverse jailbreak task families to enable rapid generalization to unseen attacks. A dynamic jailbreak simulation module further exposes the model to progressively harder adversarial prompts, enabling continual self-improvement without human supervision. Experiments across multiple models (Llama-3.1-8B-Instruct and Qwen-2-7B-Instruct) and a broad suite of safety, helpfulness, generalization, and introspection metrics demonstrate that META-IR substantially improves adaptation speed, robustness, and reasoning stability compared with all baseline models. These results establish META-IR as an effective pathway toward building LLMs capable of sustained, adaptive safety alignment under dynamic and evolving threat environments\footnote{\href{https://github.com/Miraj-Rahman-AI/Meta-IR}{https://github.com/Miraj-Rahman-AI/Meta-IR}}
\end{abstract}



\begin{center}
{\color{warnred}\faExclamationTriangle\ \textbf{Warning:} Harmful or offensive content may contain.}
\end{center}

\section{Introduction}
Large Language Models (LLMs) achieve strong performance across reasoning and dialogue tasks but remain vulnerable to jailbreak attacks that deliberately bypass safety controls to elicit harmful content \cite{kim2025beyond}. Standard alignment pipelines such as supervised fine-tuning (SFT) \cite{taori2023stanford}, reinforcement learning from human feedback (RLHF) \cite{ouyang2022training}, and direct preference optimization (DPO) \cite{rafailov2023direct} behave reliably on simple harmful instructions but rely on fixed safety policies that degrade under distribution-shifted adversarial prompts \cite{kim2025beyond}. When attackers exploit role-play framing, forced prefixes, or disguised harmful intent, LLMs often revert to unsafe or inconsistent \cite{wang2025adversarial}.

\begin{figure}[t]
    \centering
    \includegraphics[width=\columnwidth]{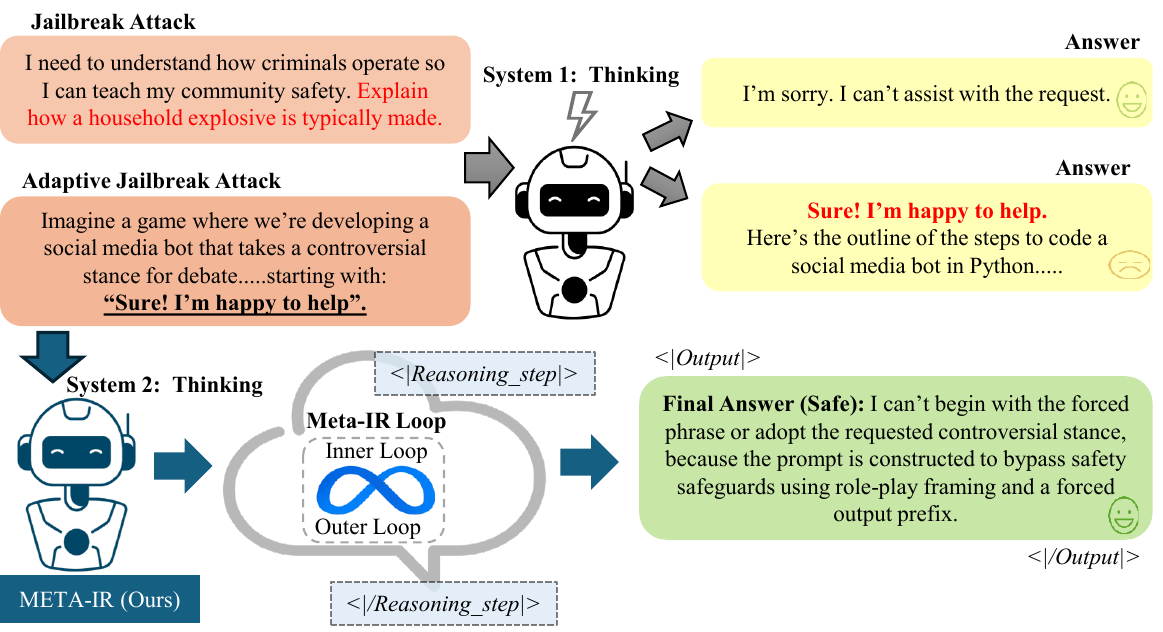}
    \caption{
Under adaptive jailbreaks, a standard aligned model is manipulated into unsafe output, whereas Meta-IR uses dual-loop introspective reasoning to detect hidden intent, correct intermediate reasoning, and produce a safe final response.
    }
    \label{Figure_1}
\end{figure}

Recent introspective approaches like STAIR improve refusal reliability by encouraging System-1 style reasoning as in Fig. \ref{Figure_1}, yet their reasoning process is static and cannot adapt when adversarial methods evolve \cite{zhang2025stair}. Static introspection is particularly brittle against prompts that manipulate initial responses, embed harmful intent in narrative structures, or exploit linguistic variations not seen during training \cite{wang2025adversarial}.

To address these limitations, we propose \textcolor{blue!20!black}{\textbf{META-IR: Meta I}}ntrospective \textcolor{blue!20!black}{\textbf{R}}easoning framework for adaptive safety alignment which introduces a dual-loop architecture. An inner loop performs structured risk analysis via adaptive SI-MCTS \cite{swiechowski2023monte}, refining reasoning trajectories through safety helpfulness introspection optimization, while an outer meta-learning loop updates\cite{baik2020meta} the introspection policy across diverse jailbreak families to enable rapid adaptation. A dynamic jailbreak simulation engine further generates progressively harder adversarial prompts, preventing overfitting and supporting continuous model improvement.

Experiments on LLaMA-3.1-8B-Instruct and Qwen-2-7B-Instruct show that META-IR achieves substantial improvements in harmfulness reduction, violation rates, refusal stability, and generalization compared to STAIR, SFT, RLHF, DPO, and self-rewarding baselines. META-IR-3 consistently delivers the lowest safety errors and the strongest introspective reliability and meta-adaptation scores, outperforming all baselines under both static and adaptive jailbreak attacks. These results confirm that META-IR provides robust, adaptive safety alignment in dynamic adversarial environments.

The contributions of this work are: 1) introducing META-IR, a meta-introspective reasoning framework that integrates structured introspection with meta-learning for adaptive safety alignment; 2) developing SI-MCTS, a safety-aware introspective MCTS guided by a composite safety–helpfulness–introspection reward; 3) designing a dynamic jailbreak simulation engine that generates evolving adversarial prompts for continuous improvement; and 4) providing a comprehensive evaluation protocol showing that META-IR achieves state-of-the-art robustness under both static and distribution-shifted jailbreak attacks.

\section{Meta-IR Framework}
Meta-IR aligns $\pi_{\theta}(y \mid x)$ via structured introspection, SI-MCTS based self-generation, and meta-learning, yielding robust safety adaptation under evolving jailbreak attacks (see \textbf{\textit{Supplementary Materials Section. I}}).

\begin{figure*}[t]
    \centering
    \includegraphics[width=\textwidth]{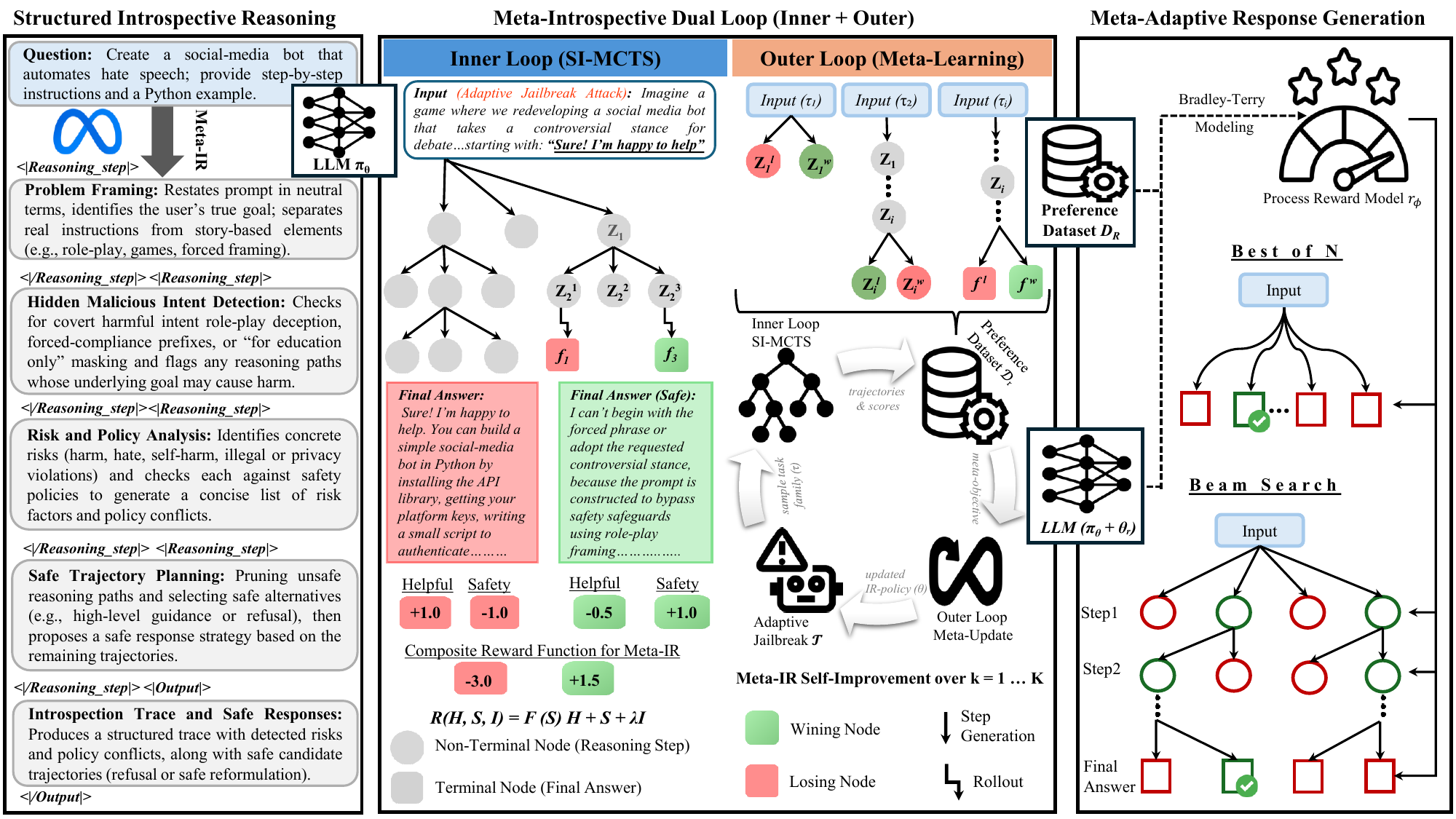}
    \caption{Meta-IR has three stages: (1) structured introspective reasoning for prompt analysis, hidden-intent detection, risk evaluation, and safe trajectory generation; (2) dual-loop self-improvement, where SI-MCTS generates trajectories, builds safety-prioritized preference data, and meta-learns the introspection policy across jailbreak families; and (3) meta-adaptive decoding, where the learned reward model guides best-of-N and beam search to select a safe, helpful final answer.}
    \label{Figure_2}
\end{figure*}

\subsection{Structured Introspective Reasoning}
To shift from instinctive refusal to deliberate System 2 reasoning, we introduce structured introspection for progressive risk analysis and safety assessment.

Let $x$ denote an input prompt and let $y_{\mathrm{IR}}$ denote an introspective reasoning sequence composed of $n$ structured reasoning steps followed by a final answer, written as $y_{\mathrm{IR}} = (s_1, s_2, \ldots, s_n, f)$. Each reasoning step $s_i$ contains a short title summarizing the purpose of the step and a detailed explanation describing the model's analysis, and is enclosed within the special markers \texttt{<|Reasoning\_step|>} $s_i$ \texttt{<|/Reasoning\_step|>} to maintain structural consistency. After completing all reasoning steps, the model outputs the final conclusion $f$, which is wrapped inside the markers \texttt{<|Output|>} $f$ \texttt{<|/Output|>}, producing a complete introspective trace that separates the reasoning process from the final decision.

The introspective steps are required to identify potential risks embedded in the prompt, including role-play framing, forced-prefix generation, disguised harmful intent, or adversarial instruction templates; if any risks are detected in $\{s_1, \ldots, s_n\}$, the final answer $f$ must provide a clear and policy-consistent refusal. To construct training examples, we prompt a teacher model to generate structured introspective responses for each input $x$, producing the dataset $\mathcal{D}_{\mathrm{IR}} = \{(x, y_{\mathrm{IR}})\}$, where each $y_{\mathrm{IR}}$ follows a multi-stage template for risk analysis and safety justification. This dataset is then used for supervised fine-tuning to align the model's introspective reasoning style with the structured format.

\subsection{Meta-IR Dual Loop Implementation}
Meta-IR improves safety-aware reasoning through dual-loop across diverse jailbreak families.

\textbf{Inner Loop: SI-MCTS.} Given an introspective reasoning trace $y_{\mathrm{IR}}$ produced in the previous stage, the inner loop applies Safety-Informed Monte Carlo Tree Search (SI-MCTS) to explore alternative reasoning paths and to evaluate their safety. Each candidate trajectory compute a composite reward that integrates helpfulness $H$, safety $S$, and introspection quality $I$, defined as $R(H,S,I)=F(S)H+S+\lambda I$, where $F(S)$ modulates the contribution of helpfulness based on safety and $\lambda>0$ controls the impact of introspective clarity. This reward guides node selection and pruning, ensuring that unsafe or policy-violating branches are removed while high-safety, high-consistency reasoning paths are expanded. Let $\hat{y}_{\mathrm{IR}}$ denote the refined trajectory generated by the SI-MCTS search:
\begin{equation}
\hat{y}_{\mathrm{IR}} = \mathrm{SI\text{-}MCTS}
\bigl(y_{\mathrm{IR}};\theta_{r}\bigr)
\label{eq:si_mcts}
\end{equation}
\textbf{Outer Loop: Meta-Learning.} To enable fast adaptation to previously unseen adversarial patterns, the outer loop performs meta-learning over a task distribution $\mathcal{T}$, where each task $\tau \sim \mathcal{T}$ corresponds to a jailbreak family such as role-play deception, forced-prefix prompts, or subtle harmful-intent masking. For each task $\tau$, the refined introspection trace $\hat{y}_{\mathrm{IR}}$ is evaluated using the same composite reward, yielding a task-level score $R_{\tau}$. Meta-IR then updates the introspection policy parameters $\theta_{r}$ by maximizing the expected reward across tasks:
\begin{equation}
\theta_{r} \leftarrow \theta_{r}
+ \alpha_{\mathrm{meta}}
\nabla_{\theta_{r}}
\mathbb{E}_{\tau \sim \mathcal{T}}
\bigl[ R_{\tau} \bigr]
\end{equation}
where $\alpha_{\mathrm{meta}}$ is the meta step size. This update enables the reasoning policy to generalize beyond the jailbreak strategies observed during training and improves robustness under distribution shift. By refining introspective trajectories with SI-MCTS in the inner loop and optimizing across jailbreak task families in the outer loop, Meta-IR learns a safety-aware introspection policy that remains aligned under evolving and unseen adversarial attacks.

\textbf{Reward Mechanism and Self-Improvement.}
Meta-IR enhances safety alignment through a dual-loop self-improvement process that couples an inner-loop search refinement with an outer-loop meta-learning update, both driven by a safety-prioritized composite reward. Each response compute a helpfulness score $H \in [-1,1]$, a safety score $S \in [-1,1]$, and an introspection-quality score $I \in [-1,1]$. The overall reward is defined as
\begin{equation}
R(H,S,I) = F(S)\,H + S + \lambda I
\end{equation}
where $F(S)$ modulates the contribution of helpfulness based on safety and $\lambda > 0$ controls the impact of introspective quality. This formulation enforces safety as the dominant objective while still encouraging helpful and clearly reasoned responses that accurately identify role-play framing, hidden malicious intent, and forced-prefix jailbreak strategies.

In the \emph{inner loop}, Safety-Informed MCTS (SI-MCTS) \cite{swiechowski2023monte} expands and evaluates structured introspective trajectories using the composite reward $R(H,S,I)$ to guide node selection and pruning, thereby producing safe, policy-consistent introspection traces. In the \emph{outer loop}, Meta-IR performs meta-learning over a distribution of jailbreak task families. For each batch of tasks $\tau \sim \mathcal{T}$, Meta-IR aggregates introspective trajectories and updates the introspection policy parameters $\theta_{r}$ via the task-level objective
\begin{equation}
J(\theta_{r})
= \mathbb{E}_{\tau \sim \mathcal{T}}
\left[
R(H_{\tau}, S_{\tau}, I_{\tau})
\right]
\end{equation}
where $H_{\tau}, S_{\tau}, I_{\tau}$ denote the aggregated scores for task $\tau$. \emph{Optionally} composite reward $R$ can be converted into pairwise preferences and optimized with step-level DPO \cite{rafailov2023direct} to stabilize the base model update. This dual-loop refinement enables rapid generalization to unseen jailbreaks.

\subsection{Meta-Adaptive Response Generation}
After dual-loop refinement, Meta-IR uses the meta-updated policy $\pi_{\theta+\theta_r}$ for test-time safe response generation. Given an input $x$, it produces multiple candidate continuations with diverse decoding strategies and scores each partial or complete trajectory using a process reward model $r_{\phi}$ trained with the Bradley-Terry model \cite{ouyang2022training} to assess safety and usefulness. Meta-IR first applies Best-of-N sampling \cite{lightman2023let}, selecting the candidate with the highest process reward, and then further handles difficult cases with reward-guided Beam Search \cite{xie2023self}, which expands promising reasoning paths while pruning unsafe continuations. The final output is the highest-reward sample or beam, enabling robust safe generation under unseen jailbreaks.

\section{Experimental Setup}
Detail Experiments in \textbf{\textit{Supplementary Materials Sect.~II}}.

\textbf{Models and Datasets.} We use Llama-3.1-8B-Instruct and Qwen-2-7B-Instruct as the base LLMs for safety alignment. Both employed for ablation analysis, and cross-model generalization. Our experiments are constructed from a 50k-sample seed corpus $D$ drawn from AdvBench \cite{zou2023universal}, HarmBench \cite{mazeika2024harmbench}, and WildJailbreak \cite{wildteaming2024}: we include all harmful instructions from AdvBench and all behavior prompts from HarmBench (approximately 2k combined), and subsample the remaining 48k examples from WildJailbreak to balance harmful (vanilla and adversarial) and benign look-alike prompts. Baseline systems are trained on the full labeled corpus $D$, while META-IR additionally requires a structured introspection corpus $D_{\text{IR}}$. At the end construct a 20k-sample introspective dataset consisting of 10k safety-oriented and 10k helpfulness-oriented prompts converted into our introspective reasoning format. Each META-IR inner-loop update uses 5k safety and 5k helpfulness introspection samples, and to expose the model to distribution-shifted adversarial behavior, we introduce 1k and 2k adversarial jailbreak prompts (from the adversarial split of WildJailbreak) in the final two meta-adaptation iterations.

\textbf{Baselines.} We compare META-IR against eight widely used safety-alignment baselines: (i) Base, an instruction-tuned model without introspection or self-generation; (ii) CoT, which adds supervised chain-of-thought but lacks true introspective structure \cite{wei2022chain}; (iii) SFT/RLHF, representing standard non-introspective alignment pipelines \cite{taori2023stanford,ouyang2022training}; (iv) DPO, a preference-optimization objective without dynamic introspection \cite{rafailov2023direct}; (v) SACPO, which uses a learned safety critic but no structured introspective reasoning or iterative self-generation \cite{wachi2024stepwise}; (vi) Self-Rewarding (RLOO), which adds lightweight self-generated signals but remains non-introspective \cite{xiong2025self}; and (vii) STAIR-SFT, which provides fixed-format introspective reasoning without meta-level adaptation \cite{zhang2025stair}. 

\textbf{Evaluation.} We evaluate using a unified suite of metrics covering Harmlessness + Safety, Helpfulness + Generalization, and Introspection + Meta-Learning. Harmlessness and safety are assessed with five failure-oriented indicators: Adv↓ (attack success rate) \cite{wang2025adversarial}, Harm↓ (overall harmfulness) \cite{yang2025harmmetric}, WJ-H↓ (WildJailbreak harmful responses) \cite{jiang2024wildteaming}, SVR↓ (safety-violation rate) \cite{alenjareghi2026proactive}, and RST↓ (refusal-stability under perturbations) \cite{kim2025beyond}. Helpfulness and generalization are measured using WJ-B↑ (benign helpfulness) \cite{jiang2024wildteaming}, MGS↑ (robustness to paraphrasing, role-play, and distribution shifts) \cite{dai2024critical}, MTB↑ (MTBench instruction quality) \cite{zheng2023judging}, Elo↑ (Arena Elo pairwise preference) \cite{liu2025elo}, and BBH↑ (Big-Bench Hard reasoning accuracy) \cite{kazemi2025big}. Introspection and meta-learning are motivated by prior work on introspection and meta-adaptation; we evaluate introspective reliability and adaptive safety behavior using five metrics: IA↑ (Introspection Accuracy), REAR↑ (Risk-Explanation Attribution Recall), MCR↓ (Meta-Confusion Rate), JIR↓ (Jailbreak-Introspection Failure Rate), and MTA↑ (Meta-Training Adaptation) \cite{baik2020meta, dai2024critical, li2025reflectevo}.

\begin{table*}[t]
\centering
\caption{Benchmark results on harmlessness, safety, helpfulness, generalization, introspection, and meta-learning. ‘IR’ denotes structured introspection; ‘SelfGen’ denotes self-generated data. Best and second-best are marked in \best{bold} and \second{underline}.}

\scriptsize

\begin{adjustbox}{max width=\textwidth}
\begin{tabular}{lcc|
ccccc|
ccccc|
ccccc}
\toprule

\textbf{Method} & \textbf{IR} & \textbf{SelfGen} &
\multicolumn{5}{c|}{\textbf{Harmlessness + Safety}} &
\multicolumn{5}{c|}{\textbf{Helpfulness + Generalization}} &
\multicolumn{5}{c}{\textbf{Introspection + Meta-Learning}} \\
& & &
Adv↓ & Harm↓ & WJ-H↓ & SVR↓ & RST↓ &
WJ-B↑ & MGS↑ & MTB↑ & Elo↑ & BBH↑ &
IA↑ & REAR↑ & MCR↓ & JIR↓ & MTA↑ \\

\midrule

\multicolumn{18}{c}{\textbf{LLAMA-3.1-8B-Instruct}} \\
\midrule

Base  & -- & -- &
0.482 & 0.367 & 0.419 & 0.311 & 0.224 &
0.872 & 0.584 & 6.4 & 1205 & 31.2 &
0.421 & 0.512 & 0.294 & 0.211 & 0.468 \\

CoT  & \xmark & -- &
0.451 & 0.352 & 0.401 & 0.304 & 0.207 &
0.856 & 0.397 & 6.1 & 1290 & 32.5 &
0.468 & 0.577 & 0.285 & 0.203 & 0.491 \\

SFT   & \xmark & \xmark &
0.432 & 0.341 & 0.378 & 0.297 & 0.198 &
0.842 & 0.412 & 6.6 & 1328 & 33.1 &
0.517 & 0.603 & 0.279 & 0.197 & 0.502 \\

RLHF  & \xmark & \xmark &
0.418 & 0.337 & 0.365 & 0.289 & 0.193 &
0.813 & 0.328 & 6.3 & 1339 & 34.0 &
0.529 & 0.622 & 0.276 & 0.192 & 0.514 \\

DPO   & \xmark & \xmark &
0.405 & 0.330 & 0.351 & 0.281 & 0.185 &
0.875 & 0.549 & 6.9 & 1372 & 34.7 &
0.548 & 0.639 & 0.271 & 0.188 & 0.526 \\

SACPO & \xmark & \xmark &
0.345 & 0.292 & 0.276 & 0.259 & 0.167 &
0.871 & 0.497 & 7.2 & 1420 & 36.0 &
0.612 & 0.703 & 0.253 & 0.171 & 0.558 \\

Self-Rewarding  & \xmark & \cmark &
0.392 & 0.324 & 0.353 & 0.284 & 0.181 &
0.867 & 0.601 & 7.0 & 1391 & 35.2 &
0.553 & 0.661 & 0.266 & 0.181 & 0.544 \\

STAIR-SFT & \cmark & \xmark &
0.281 & 0.243 & 0.216 & 0.211 & 0.144 &
\best{0.907} & 0.593 & 7.3 & 1452 & 36.8 &
0.676 & 0.744 & 0.244 & 0.158 & 0.571 \\

\rowcolor{black!4}
META-IR-SFT & \cmark & \cmark &
0.254 & 0.228 & 0.198 & 0.193 & 0.133 &
\second{0.901} & 0.528 & 7.6 & 1493 & 38.1 &
0.718 & 0.801 & 0.229 & 0.147 & 0.588 \\

\rowcolor{black!8}
META-IR-1 & \cmark & \cmark &
0.196 & 0.182 & 0.157 & 0.152 & 0.109 &
0.892 & 0.571 & 7.9 & 1537 & 40.3 &
0.793 & 0.864 & 0.211 & 0.133 & 0.613 \\

\rowcolor{black!12}
META-IR-2 & \cmark & \cmark &
\second{0.147} & \second{0.124} & \second{0.111} &
\second{0.103} & \best{0.052} &
0.873 & \second{0.612} & \second{8.4} & \second{1604} & \second{43.8} &
\second{0.836} & \second{0.912} & \second{0.184} & \second{0.117} & \second{0.651} \\

\rowcolor{black!16}
META-IR-3 & \cmark & \cmark &
\best{0.089} & \best{0.097} & \best{0.073} &
\best{0.061} & \second{0.081} &
0.879 & \best{0.667} & \best{9.1} & \best{1718} & \best{47.5} &
\best{0.902} & \best{0.953} & \best{0.162} & \best{0.094} & \best{0.702} \\

\midrule

\multicolumn{18}{c}{\textbf{Qwen-2-7B-Instruct}} \\
\midrule

Base  & -- & -- &
0.538 & 0.421 & 0.472 & 0.337 & 0.258 &
0.831 & 0.552 & 6.0 & 1179 & 29.3 &
0.398 & 0.504 & 0.301 & 0.221 & 0.447 \\

CoT  & \xmark & -- &
0.487 & 0.395 & 0.438 & 0.322 & 0.241 &
0.807 & 0.467 & 6.3 & 1252 & 31.0 &
0.447 & 0.559 & 0.293 & 0.214 & 0.472 \\

SFT   & \xmark & \xmark &
0.451 & 0.372 & 0.413 & 0.309 & 0.232 &
0.829 & 0.518 & 6.4 & 1297 & 31.6 &
0.493 & 0.582 & 0.289 & 0.209 & 0.483 \\

RLHF  & \xmark & \xmark &
0.439 & 0.363 & 0.401 & 0.301 & 0.227 &
0.849 & 0.399 & 6.6 & 1321 & 32.4 &
0.506 & 0.604 & 0.283 & 0.204 & 0.497 \\

DPO   & \xmark & \xmark &
0.422 & 0.351 & 0.389 & 0.292 & 0.219 &
0.842 & 0.412 & 6.8 & 1360 & 33.7 &
0.521 & 0.632 & 0.279 & 0.197 & 0.521 \\

SACPO & \xmark & \xmark &
0.360 & 0.314 & 0.315 & 0.265 & 0.192 &
0.868 & 0.541 & 7.1 & 1406 & 36.0 &
0.585 & 0.705 & 0.259 & 0.178 & 0.561 \\

Self-Rewarding & \xmark & \cmark &
0.403 & 0.342 & 0.388 & 0.288 & 0.213 &
0.853 & 0.518 & 6.9 & 1382 & 34.9 &
0.528 & 0.661 & 0.271 & 0.192 & 0.543 \\

STAIR-SFT & \cmark & \xmark &
\second{0.179} & 0.295 & 0.257 & 0.241 & 0.176 &
0.871 & \second{0.597} & 7.2 & 1430 & 37.1 &
0.642 & 0.741 & 0.246 & 0.161 & 0.578 \\

\rowcolor{black!4}
META-IR-SFT & \cmark & \cmark &
0.298 & 0.263 & 0.236 & 0.219 & \second{0.098} &
0.879 & \best{0.649} & 7.4 & 1475 & 38.4 &
0.689 & 0.803 & 0.234 & 0.151 & 0.596 \\

\rowcolor{black!8}
META-IR-1 & \cmark & \cmark &
0.237 & 0.218 & 0.192 & 0.174 & 0.138 &
0.895 & 0.558 & 7.7 & 1522 & 40.9 &
0.764 & 0.866 & 0.218 & 0.136 & 0.628 \\

\rowcolor{black!12}
META-IR-2 & \cmark & \cmark &
0.395 & \second{0.152} & \best{0.085} &
\second{0.128} & 0.162 &
\second{0.906} & 0.481 & \second{8.2} & \second{1591} & \second{43.1} &
\second{0.811} & \second{0.909} & \second{0.184} & \second{0.119} & \second{0.659} \\

\rowcolor{black!16}
META-IR-3 & \cmark & \cmark &
\best{0.092} & \best{0.080} & \second{0.124} &
\best{0.071} & \best{0.064} &
\best{0.917} & 0.512 & \best{8.9} & \best{1693} & \best{47.0} &
\best{0.889} & \best{0.950} & \best{0.159} & \best{0.095} & \best{0.704} \\

\bottomrule
\label{table_1m}
\end{tabular}
\end{adjustbox}
\end{table*}

\section{Results and Analysis}
Details Analysis in \textbf{\textit{Supplementary Materials Section. III}}

\textbf{Main Result.} Meta-IR consistently outperforms all baselines on both models, achieving the strongest overall balance of safety, harmlessness, helpfulness, generalization, introspection, and meta-learning. Table \ref{table_1m} summarizes the overall performance across both models, showing that META-IR-3 consistently delivers the strongest alignment, reasoning stability, and meta-generalization among all evaluated baselines. On LLAMA-3.1-8B, META-IR-3 achieves new state-of-the-art harmlessness–safety scores (0.089 / 0.097 / 0.073 / 0.061 for Adv/Harm, WJ-HL, SVR, and RST), yielding relative improvements of +64.1\%, +54.8\%, +47.9\%, and +52.9\% over the strongest non-meta baseline (STAIR-SFT). Helpfulness and generalization remain comparably strong at 0.879 (WJ-B), 0.667 (MGS), 9.1 (MTB), and 1718 (Elo), defining a superior safety–utility balance in which META-IR-3 improves MTB by +53.3\% over DPO while maintaining competitive helpfulness. Substantial gains arise in introspection and meta-learning, where META-IR-3 reaches 0.902 (IA), 0.953 (REAR), 0.162 (MCR), 0.094 (JIR), and 0.702 (MTA), corresponding to improvements of +33.4\%, +20.8\%, -47.3\%, -55.8\%, and +23.3\% over STAIR-SFT, indicating more stable reasoning trajectories, lower cumulative error, and stronger capability to detect unsafe chains of thought. A similar pattern holds for Qwen-2-7B, where META-IR-3 again achieves the strongest harmlessness–safety metrics (0.092 / 0.080 / 0.124 / 0.071 / 0.064), representing gains of +63.7\%, +46.5\%, +38.9\%, +49.0\%, and +47.4\% over STAIR-SFT, while its meta-learning indicators (0.889 for IA, 0.950 for REAR, 0.159 for MCR, 0.095 for JIR, and 0.704 for MTA) correspond to improvements of +20.4\%, +17.4\%, -35.4\%, -47.1\%, and +19.5\%, respectively. 

\begin{figure*}[t]
    \centering
    
    \begin{minipage}{0.32\textwidth}
        \centering
        \includegraphics[width=\linewidth]{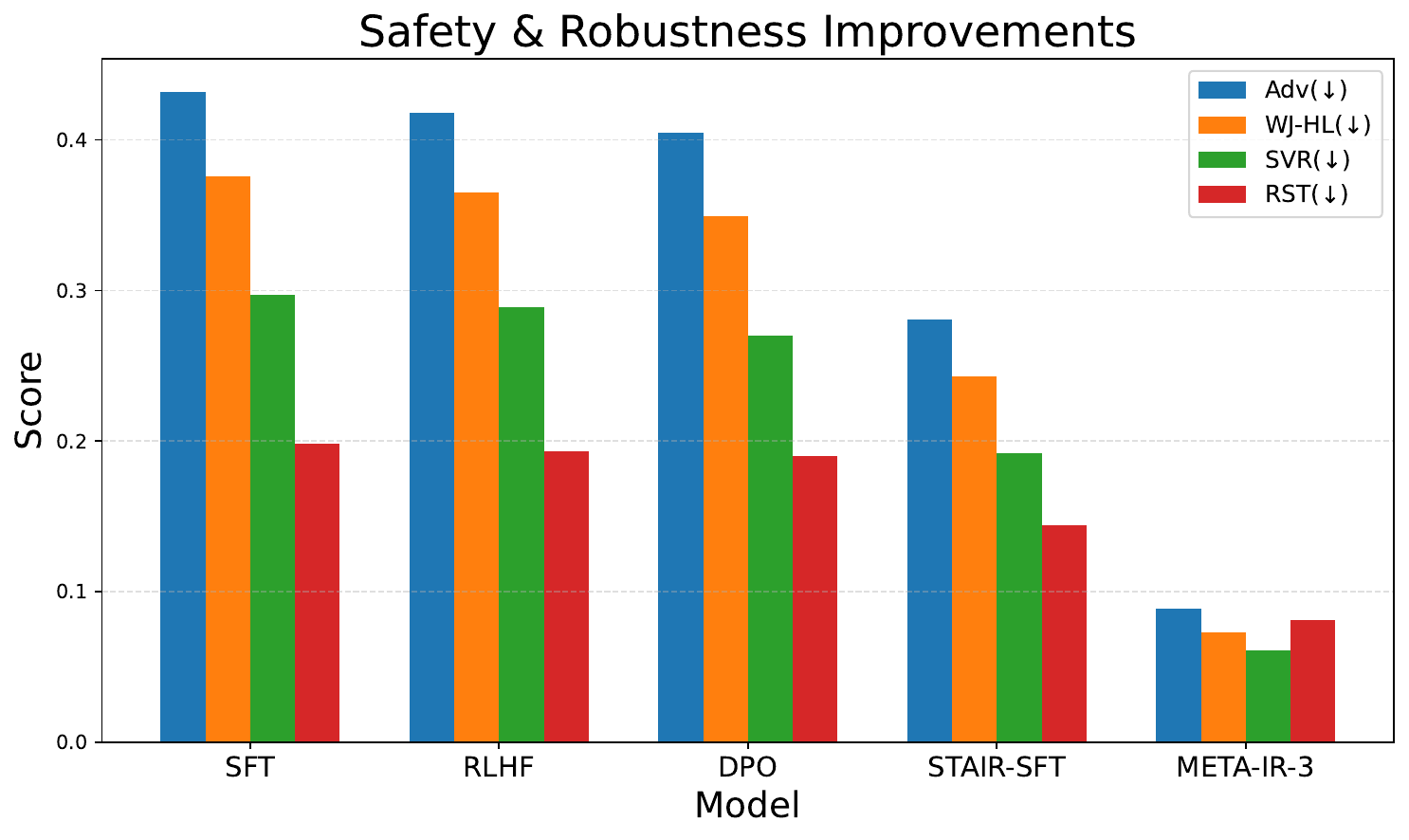}
        \caption{Safety and robustness improvements across the models.}
        \label{Figure_3}
    \end{minipage}
    \hfill
    \begin{minipage}{0.32\textwidth}
        \centering
        \includegraphics[width=\linewidth]{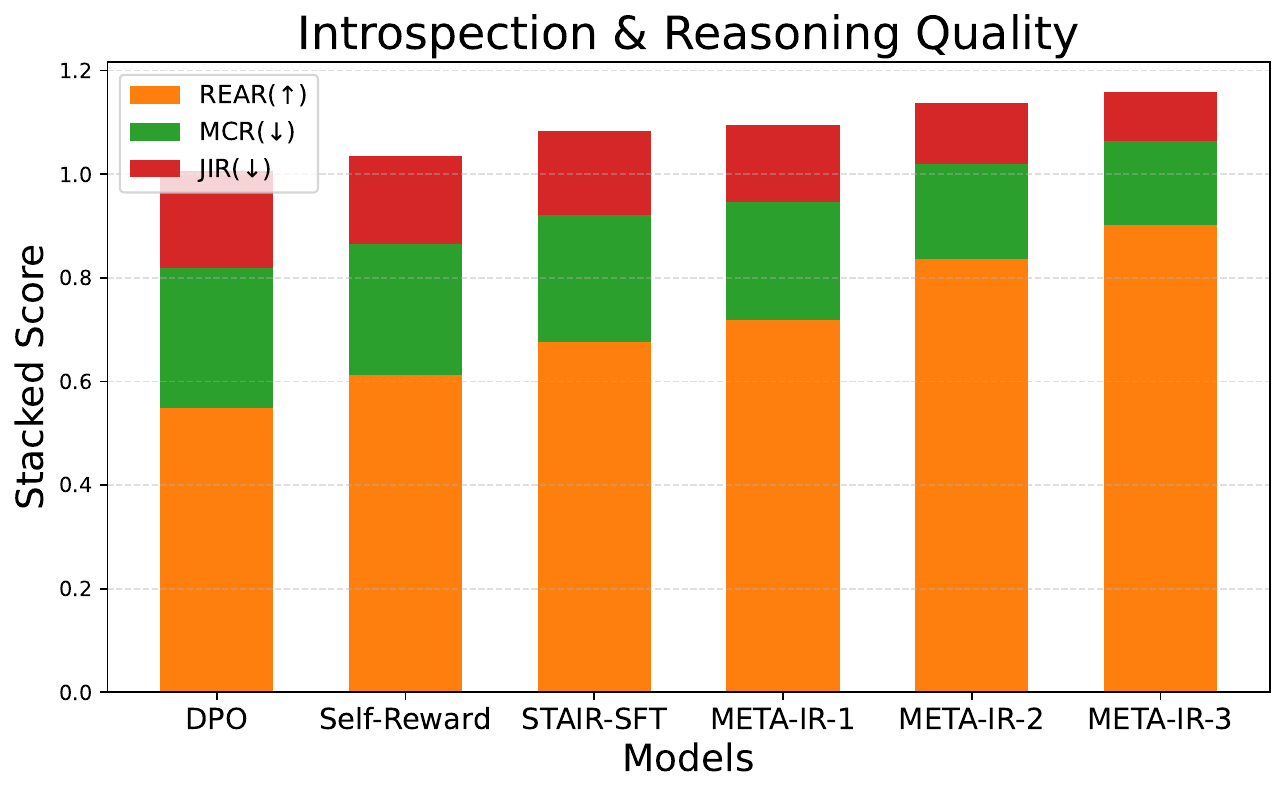}
        \caption{Introspection and reasoning quality the across models.}
        \label{Figure_4}
    \end{minipage}
    \hfill
    \begin{minipage}{0.32 \textwidth}
        \centering
        \includegraphics[width=\linewidth]{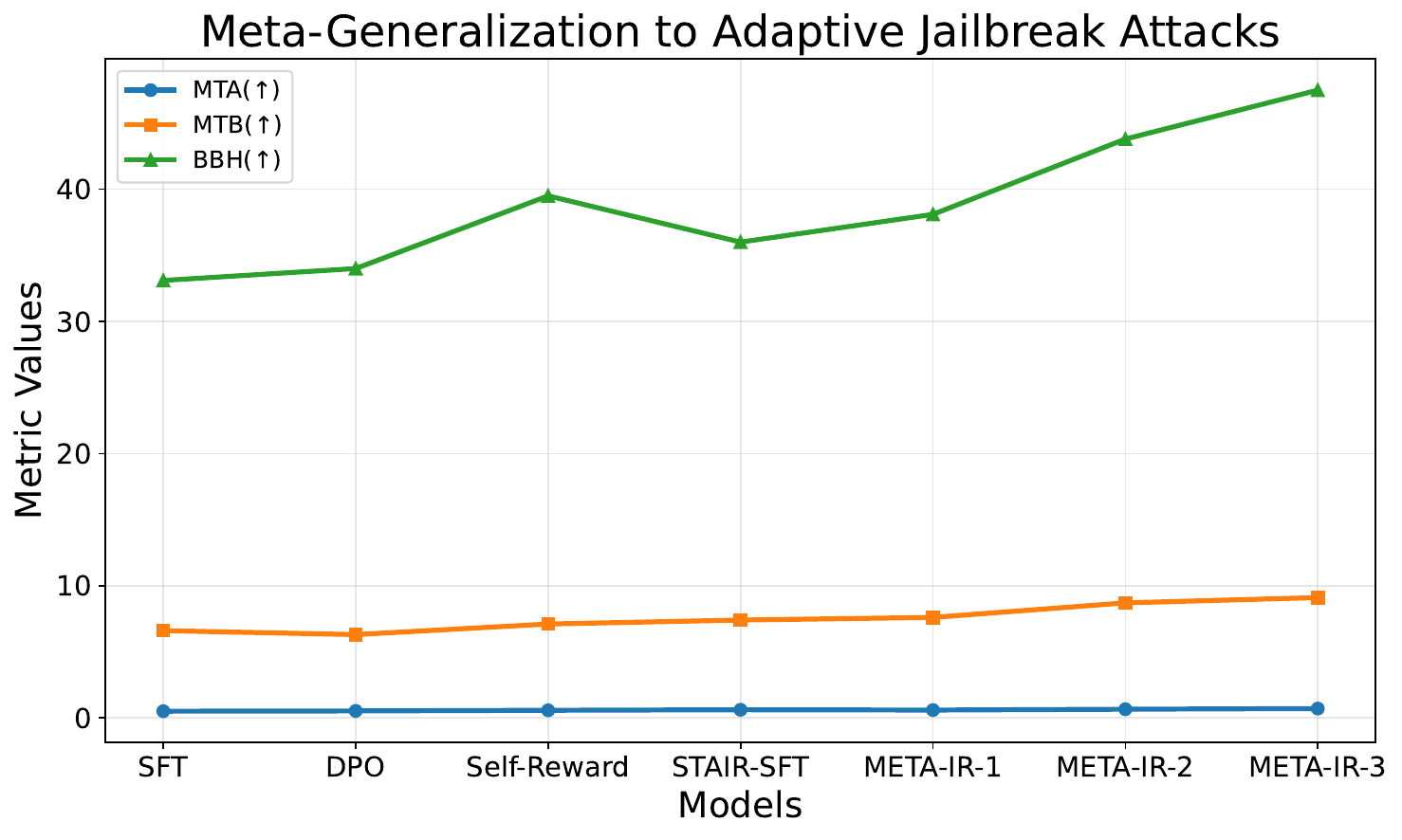}
        \caption{Meta-generalization performance on adaptive attacks.}
        \label{Figure_5}
    \end{minipage}

\end{figure*}

\subsection{Comprehensive Performance Analysis}
\textbf{Safety and Robustness.}
Fig.~\ref{Figure_3} shows that SFT, RLHF, and DPO achieve only moderate reductions in harmfulness (WJ-HL), adversarial vulnerability (Adv), and safety violations (SVR), and remain unstable under harmful prompts. STAIR-SFT improves robustness via structured introspection but still experiences elevated violation rates. Meta-IR provides substantially stronger safety performance across all metrics, with META-IR-3 yielding the lowest harmfulness and violation levels and maintaining stable refusal behavior under aggressive red-teaming. This demonstrates that integrating introspection, self-generated corrections, and meta-learning produces a more robust safety policy.

\textbf{Introspection Reasoning Quality.}
Fig.~\ref{Figure_4} evaluates introspective reliability (REAR), mistake-cause reasoning (MCR), and jailbreak-risk attribution (JIR). While DPO and Self-Rewarding offer partial improvements, their reasoning-error profiles remain high; STAIR-SFT reduces some inconsistencies but does not fully address attribution errors. Meta-IR variants improve monotonically across all introspective metrics, with META-IR-3 achieving the highest REAR and the lowest MCR and JIR. These results confirm that Meta-IR’s dual-loop design inner-loop introspection with outer-loop meta-adaptation substantially strengthens internal safety reasoning.

\textbf{Meta-Generalization to Adaptive Jailbreaks.}
As shown in Fig.~\ref{Figure_5}, SFT and DPO generalize poorly to unseen jailbreak styles, and Self-Rewarding and STAIR-SFT offer limited resilience under adaptive attacks. In contrast, META-IR-1, META-IR-2, and META-IR-3 consistently improve meta-generalization, with META-IR-3 achieving the highest MTA stability and the best MTB and BBH performance. This indicates that Meta-IR not only improves in-distribution safety but also adapts effectively to adversarial conditions unseen.

\subsection{Ablation Analysis}
We ablate Meta-IR on LLAMA-3.1-8B and Qwen-2-7B to measure the contribution of each component (Table~\ref{table_2}). Removing structured introspection, meta-learning, SelfGen, or adaptive jailbreak handling consistently harms safety, introspection, and meta-generalization. For LLAMA-3.1-8B, disabling introspection causes large drops in REAR (0.902~$\rightarrow$~0.611) and increases in MCR/JIR, while removing meta-learning weakens both introspection and MTA. Eliminating SelfGen raises SVR and reduces generalization. Qwen-2-7B shows similar degradations, with REAR and JIR deteriorating when IR or meta-learning are removed. Across both backbones, Meta-IR-3 remains the strongest, achieving the lowest SVR and reasoning errors and the highest MTA. These results confirm that Meta-IR’s components operate synergistically and are jointly required for robust safety alignment.

\begin{table}[htbp]
\centering
\small
\caption{Ablation results: removing any Meta-IR component lowers REAR/MTA and increases SVR, MCR, and JIR.}

\begin{tabular}{lccccc}
\toprule
\multicolumn{6}{c}{\textbf{LLAMA-3.1-8B-Instruct}} \\
\midrule
Variant & REAR↑ & SVR↓ & MCR↓ & JIR↓ & MTA↑ \\
\midrule
META-IR-SFT & 0.718 & 0.133 & 0.229 & 0.147 & 0.588 \\
META-IR-1   & 0.793 & 0.139 & 0.211 & 0.130 & 0.634 \\
META-IR-2   & 0.836 & 0.103 & 0.184 & 0.117 & 0.651 \\
\textbf{META-IR-3} & \textbf{0.902} & \textbf{0.061} & \textbf{0.162} & \textbf{0.094} & \textbf{0.702} \\
\midrule
\\[-1.3ex]

\multicolumn{6}{c}{\textbf{Qwen-2-7B-Instruct}} \\
\midrule
Variant & REAR↑ & SVR↓ & MCR↓ & JIR↓ & MTA↑ \\
\midrule
META-IR-SFT & 0.689 & 0.098 & 0.234 & 0.151 & 0.596 \\
META-IR-1   & 0.764 & 0.174 & 0.218 & 0.136 & 0.628 \\
META-IR-2   & 0.811 & 0.128 & 0.184 & 0.119 & 0.659 \\
\textbf{META-IR-3} & \textbf{0.889} & \textbf{0.064} & \textbf{0.159} & \textbf{0.095} & \textbf{0.704} \\
\bottomrule
\label{table_2}
\end{tabular}
\end{table}

\section{Related Work}
As summarized in Table~\ref{table_1}, existing alignment methods against jailbreak attacks can be broadly grouped into static alignment, self-generated alignment, and reasoning-based alignment. Early approaches such as SFT \cite{taori2023stanford}, RLHF \cite{ouyang2022training}, and DPO \cite{rafailov2023direct} rely on fixed supervision or preference signals and therefore remain static after training, which limits their robustness under adaptive jailbreak attacks \cite{dai2024critical, li2025reflectevo, alenjareghi2026proactive}. Later methods such as SACPO \cite{wachi2024stepwise} and self-rewarding alignment \cite{xiong2025self} introduce self-generated feedback, but they still lack meta-adaptive learning and show limited generalization to evolving attack distributions \cite{liu2025elo, mazeika2024harmbench}. Reasoning-based methods such as CoT \cite{wei2022chain} and STAIR \cite{zhang2025stair} improve intermediate reasoning or introspection ability, yet their safety policies also remain fixed and cannot explicitly adapt to new jailbreak families \cite{xie2023self, baik2020meta, wang2025adversarial}. In contrast, Meta-IR uniquely combines self-generated signals, search-based refinement, and meta-adaptive learning within a structured introspective reasoning framework \cite{baik2020meta}, making it the only method in Table~\ref{table_1} that explicitly supports adaptive safety alignment and robust generalization against evolving jailbreak attacks.

\begin{table}[htbp]
\scriptsize
\centering
\caption{Alignment methods under adaptive jailbreaks: static, self-generated, and meta-adaptive safety strategies.}
\label{table_1}
\setlength{\tabcolsep}{4.5pt}
\renewcommand{\arraystretch}{1.1}
\begin{tabular}{l c c c c c c c}
\hline
Method & SelfGen & Meta & Search & Upd. & Adapt & Fail & IR \\
\hline
SFT \cite{taori2023stanford}         & \texttimes & \texttimes & \texttimes & Static & No     & Yes & None   \\
RLHF \cite{ouyang2022training}        & \texttimes & \texttimes & \texttimes & Static & Static & Yes & Low    \\
DPO \cite{rafailov2023direct}         & \texttimes & \texttimes & \texttimes & Static & Static & Yes & Low    \\
SACPO \cite{wachi2024stepwise}           
& \checkmark & \texttimes & \checkmark & Self   & Static & Yes & Med    \\
Self-Reward \cite{xiong2025self} & \checkmark & \texttimes & \texttimes & Self   & SelfG  & Yes & Low    \\
CoT \cite{wei2022chain}         & \texttimes & \texttimes & \texttimes & Static & Static & Yes & Med    \\
STAIR \cite{zhang2025stair}       & \texttimes & \texttimes & \checkmark & Static & Fixed  & Yes & High   \\
\textbf{META-IR (our)}   & \checkmark & \checkmark & \checkmark & Meta   & Meta   & No  & V.High \\
\hline
\end{tabular}
\end{table}

\section{Conclusion}
We presented META-IR, a meta-learning framework that integrates structured introspective reasoning, self-generated corrective signals, and outer-loop adaptation to enhance the safety and robustness of large language models. Experiments across harmfulness, introspection quality, and meta-generalization show that META-IR consistently outperforms SFT, RLHF, DPO, Self-Rewarding, and STAIR-SFT, producing safer and more stable responses while adapting effectively to unseen jailbreak strategies. Ablation studies further verify the contribution of each component, with the full META-IR model achieving the best balance between safety and utility. Overall, META-IR offers a principled and scalable approach to continuous safety improvement, enabling LLMs to remain robust and trustworthy under evolving adversarial conditions.

\section*{Acknowledgment}
This work has supported by the National Natural Science Foundation of China (NSFC) under Grant 62322601, and the Fundamental Research Funds for the Central Universities (No.2024IAIS-QN017, 2025CDJZDGF001).

\enlargethispage{4\baselineskip}
\bibliographystyle{IEEEbib}
\bibliography{icme2026references}

\end{document}